\documentclass{llncs}

\usepackage{array} 
\usepackage{enumerate}
\usepackage{amssymb,xypic,latexsym,graphicx}

\begin{document}

\title{A geometry of information, I: Nerves, posets and differential forms.}
\author{Jonathan Gratus\inst{1} \and Timothy Porter\inst{2}}

\institute{Department of Mathematics, School of Informatics,\\
  University of Wales Bangor, Bangor, Gwynedd, LL57 1UT, Wales, U.K.\\
\email{j@gratus.net}
\and
  Department of Mathematics, School of Informatics,\\
  University of Wales Bangor, Bangor, Gwynedd, LL57 1UT, Wales, U.K.\\
\email{t.porter@bangor.ac.uk}}

\date{\today}
\maketitle

\begin{abstract}
The main theme of this workshop (Dagstuhl seminar 04351) is `Spatial Representation: Continuous vs. Discrete'.  Spatial representation has two contrasting but interacting aspects (i) representation \emph{of} spaces' and (ii) representation \emph{by} spaces. In this paper\footnote{This paper, and the corresponding part II, evolved from the talks entitled \emph{ Fractafolds, their geometry and topology: a test bed for spatial representation}, given at the Seminar, as a result of the insights gleaned by the authors during the excellent sessions of the week. This paper with some more figures is also available on their website: http://drops.dagstuhl.de/portals/04351/ .} we will examine two aspects that are common to both interpretations of the theme, namely nerve constructions and refinement.  Representations change, data changes, spaces change. We will examine the possibility of a `differential geometry' of spatial representations of both types, and in the sequel give an algebra of differential forms that has the potential to handle the dynamical aspect of such a geometry.  We will discuss briefly a conjectured class of spaces, generalising the Cantor set which would seem ideal as a test-bed for the set of tools we are developing.
\footnote{This research forms part of a project : \emph{Fractafolds, their geometry and topology}, partially supported by a grant from the Leverhulme Trust.  This help is gratefully acknowledged.  }

\end{abstract}

\section{Introduction \label{intro}}

Spatial representation has two contrasting but closely related  aspects: (i) representation \emph{of} spaces and (ii) representation \emph{by} spaces.  The first is, classically, based firmly in geometry, and topology   and assumes some `space' is given, whilst its aim is to study the `attributes' of the space - essentially its geometry and topology, or more precisely those parts that are amenable to study by the usual tools of geometry and topology!  The other aspect represents some configuration by a space.  This `configuration' may be a formal situation modelling some relationship between some objects and attributes, or perhaps a physical context such as the space of physical configurations  of  a molecule.

In this paper and its sequel, \cite{JGTP:2004b}, we will examine two ideas shared by these different views of spatial representation: nerve constructions and refinements, thus leading to ideas of approximating the `idealised ' space by constructive and more computationally amenable gadgets, namely simplicial complexes and related posets.  As we will see, this process of approximation raises almost philosophical problems about the nature of spaces and how we observe them. The two aspects of spatial representation with which we started are mirrored here by two views of what is  being done.  In the first more classical one,  the given space is approximated by the simplicial complexes, whilst for the second view, the situation is more that the `space' is made up of idealised `points' that are observable via the approximating process and the ultimate consequence of that view is that the `space' \emph{is} the approximating system of simpler gadgets and thus the notion of `point' becomes itself an idealisation.   Spaces are rarely `static'.  The space may change or we may be looking at a changing `point' within the space.  A challenging problem is thus to study change in the context of spatial representations.  If a state of a system is represented by a `point' of a space, it is natural to seek for the evolution of that state under some `dynamics' given perhaps by some `energy' functions, vector fields, etc. The usual classical expedient is to \emph{impose} the structure of a smooth manifold on the state space, but in a system based on `observations' , \emph{the state space cannot be observed to be a manifold}.  One is left with a situation in which, for the usual theory to apply, one needs smoothness, but that is computationally infeasible.  This sort of problem is at the heart of the problem of reconciling general relativity and quantum mechanics, but is also a problem, here, in modelling processes in computer science and AI.  One needs a discrete analogue of differential geometry which behaves well with regard to the viewpoint of approximations. This type of theory has been under development in mathematical physics (e.g. \cite{mallios-raptis:Cech-deRham:2001}), but has not been examined for its applicability to  areas where Information changes.  There are obvious analogies. This will give us a second theme for this paper.

One of the approximation tools we will use is that of the nerve.  Various nerve constructions are widely used in mathematics and computer science (including AI).  They have also served as a link between continuous and discrete models in other contexts, for instance, relating to economics (even explicitly, cf. Baryshnikov, \cite{bar:1997}). They give, at any `resolution' a simplicial complex and behave moderately well under refinement of the resolution.

The title of the talks at the workshop contained the word `fractafold'. What are they and how are they relevant to this theme?  In this research, we are investigating a class of idealised spaces that could be used as a testbed for any emerging analogue of the tools of differential geometry in the observational context.  These spaces should include some well known ones such as the Cantor set and less well known ones such as the solenoids and Menger manifolds, (for which see later in this paper).  These example spaces have various common features including reasonably constructive local properties, specified by local iterated function schemes.  They are fractal, but not too wildly so! They should provide a controlled environment for testing the emergent theory, but better than that, certain of their aspects suggest that they model a class of very computationally interesting spaces.

In more detail:

\textbf{Represention of spaces?}

Spatial phenomena, spatial data, comes in many flavours.  A  differential geometer may say `Given $M$ a smooth manifold', but how is $M$ \emph{given} to us?  `Let $X$ be a space having the homotopy type of a CW complex...' says an algebraic topologist.  A worker in fractal geometry may specify a space by an iterated function scheme (IFS) or as an attractor of a dynamical system. An algebraic geometer may specify a geometric object by some polynomial equations or as a moduli space.

Outside mathematics as such, a medical imager may have  point cloud data output from a scanner; a digital topologist has a digitalised pixelled image or  some volume data. In both cases, the discrete data is approximating some idealised spatial model. One `flavour' of such data is however often incompatible with another.  Give an algebraic topologist a manifold or CW complex and an analysis of certain invariants can be attempted, give that same algebraic topologist an algebraic curve or some  strange attractor of a dynamical system and ...! (An excellent commentary on mathematical aspects of spaces, points, etc.  can be found in Cartier's beautiful IHES birthday article, \cite{cartier}.)

\medskip
\textbf{Represention by spaces}

In AI and in various other parts of Computer Science, one finds numerous instances of representation \textbf{by} spaces, that is, representation of non-spatial information in a spatial form to aid visualisation, or to use spatial analogies to help the understanding of the data.  For instance, in problems of knowledge representation there is a considerable literature on analyses of relational data of the following form: we have a set, $\mathcal{O}$ of objects and a set, $\mathcal{A}$ of attributes and a relation, $\mathcal{R}$ from $\mathcal{O}$ to $\mathcal{A}$ modelling the idea that $o\mathcal{R}a$ means that object $o$ has attribute $a$. From this basic situation, one type of analysis yields a concept lattice, as in the talks by Pfaltz and Zhang in this workshop, but another gives a simplicial complex that represents the relation faithfully and allows various aspects of the situation to be extracted quite efficiently. (We will look at this in slightly more detail later.)  There is a sense in which the set of objects is being structured via the \emph{attributes as observations}.  Two objects may be distinguished if one of the attributes can yields different results on them, i.e., one satisfies the attribute, the other does not. Likewise, two objects are in this context equivalent if they satisfy exactly the same attributes. The simplicial complexes that result are instances of nerve constructions, and refinement of the attributes, by adding in new ones, correspond to simplicial mappings between the corresponding nerves. 

Topological spaces are used extensively in theoretical physics and increasingly in ecomonics and social choice theory (cf. \cite{bar:1993,bar:1997,chichilnisky} and other work by the same authors and their collaborators).  Some of this work is combinatorially based using simplicial complexes or simplicial sets as the principal tool, with spaces occuring as a backdrop  via some notion of triangulation.  This combination of intuitively defined configuration space managed by a combinatorial model is frequent.

Other objects such as partially ordered sets, domains, lattices, locales, profinite spaces, etc., are currently used for the study of various logical, semantical and informational situations.  Some of these have obvious topological content, others less so.  A good example of a logical situation in which intuitions from algebraic topology naturally occur is in the semantics suggested for the proof theory of the intuitionistic form of the modal logic $S4$ (Goubault and Goubault-Larrecq, \cite{EGJG:2003}), in which various forms of semantics are shown to be consequences of one based on simplicial sets, a near relative of the combinatorial gadgets, simplicial complexes, that we will be using extensively to give discrete approximations to `spaces'. Another example is the second author's attempt, \cite{TP:GAMAS}, to understand models for multiagent systems using homotopical (and thus topological) ideas based on sets with several equivalence relations.  Finally the study of distributed systems and some of the associated problems, especially that of detection of deadlock, have resulted in enriched topological models being used as an aid in the design of the algorithms, see the talks by Fajstrup and Raussen on directed homotopy in this workshop and the papers from the GETCO conferences and workshops, e.g. \cite{LFEGMR}. 

\textbf{Multiscaling, sampling and observations.}

In many of the spatial contexts mentioned above, a `space' is `observed'.  Data about the `points' of the space are calculated, or collected, then sampled and used to make a `model' of the space.  Different levels of observation may be used.  The methods used are simplicial with density of sampling sometimes being determined by the local geometry of the `space', cf. the excellent notes by Dey, \cite{TD}.

In all of these cases, the `spaces' may also be evolving in time, so different `snapshots' of the resulting `space-time' may be used.  

Combining these aspects we can extract  certain features, although not all are present in all examples:
\begin{itemize}
\item \textbf{pointless} : although a useful `sham', points are never observed; two `points' in a space can sometimes be distinguished if an observation gives significantly different values at these two sites, but can never be `equal', (a useful perspective on this type of viewpoint can be found in Vickers, \cite{Vickers}).  This suggests that `pointless' models such as locales and quantales, etc. may eventually be needed. `Open sets' are somehow more basic than `points' from this point of view.
\item \textbf{relational} : if at one resolution / scale / magnification, two points cannot be distinguished, but at another, they can be, then this tells one about their `nearness', (but, of course, this does not necessarily involve distance).
\item \textbf{directional}: one sample is a subset of another; one set of observations is a refinement of another; time is not reversible.
\item \textbf{dynamical} : when a `space' represents a situation and that situation evolves under `external' influences then the representations must also evolve.  If a state is observed in a `state space', and there are influences that cause a change of state, the observations should be able to detect and to study those changes.
\item \textbf{multiscale} : to combine fine detail with local structure.
\end{itemize}
This suggests that it would be of considerable use to have a way of handling `flows on a space', `gradient vector fields' perhaps `Morse theory' (which is related to the calculus of variations on the space), as observed by our phantom observer, i.e., as mirrored in the approximating system of `constructive' spaces.
(It is important to note that some parts of such a list of analogues for `differential' tools do exist for simplicial complexes, but each such complex encountered in this theory will be merely a snapshot of the overall situation and our hope is to extend from any one approximation to the idealised `limit' space.  Once again, how do the approximating spaces reflect change of the space or change on the space.

\textbf{Remark}

The ubiquity of this sort of situation from a categorical viewpoint was the main theme of \cite{JMCTP} and some attempts to assess the relevance of such approximations to general problems in AI was made in \cite{TP:formal:1994,TP:greylevel:1994}.

\textbf{Example - with abstract data}

The Menger cube is a higher dimensional analogue of the Cantor set. It can be imagined via its approximations and from a computational or contructive viewpoint is clearly an idealisation of a process.  The pictures of it can easily be found by searching the internet. The `space' can be precisely specified as an intersection of approximating polyhedra.  The IFS / fractal structure gives relationships between each level and the next.  The `points' of the space are ideal entities, abstracted from the approximating system that `specifies' or perhaps `defines' the space.  Observationally the `space' \emph{is} its approximating system.  Its regularity makes this a useful `toy model'.  Menger manifolds, \cite{Chig}, combine this locally well behaved fractal structure with global geometric structure.

\textbf{Fractafolds, a test bed for spatial representation tools?}

Is there some class of spaces containing the Menger manifolds, for instance, that combine sufficient `regularity', so as to enable global structure not to be dominated by local irregularities, and yet are also reasonably `irregular' in having lots of local singularities?  Some `fractal' analogue of the manifold concept might be feasible, yet it is not certain that Menger manifolds are general enough to be optimal.  In any case such a class would provide a well behaved `test bed' for new techniques to analyse global geometric structure of even more general spatial phenomena.

Such an idea has been floated by Strichartz, \cite{strichartz}, who coined the term `fractafold', and that idea relates well to Kigami's analysis on fractals, \cite{kigami}, (see also \cite{yamaguchi}), but for what we want, Kigami's theory does not seem quite right as no detailed geometric theory seems to be available there.  For global `homotopy structure', shape theory, \cite{Bor,JMCTP,MarSeg}, and its stronger relative, \cite{Mar}, can be useful. Some of the techniques from there may be adapted to a more geometric context.  Topologically, various classes of spaces may be candidates for being subclasses of some class of `fractafolds'.  These would include the solenoids, and the Menger manifolds, \cite{Chig}, the latter being spaces locally like the Menger cube.  The test to be applied is whether or not global \emph{geometric} structure can be analysed using analogues of (differential) geometric concepts  such as curvature, torsion, differential forms, vector fields, characteristic classes, etc.  and, from the perspective of this paper, how would this interpret in terms of a dynamic theory of observable change of spatial representations.

When representing a situation spatially, the tradition has been to assume local niceness conditions (manifold, CW-complex, etc.), but how can we know that such local conditions are valid.  It may be that it was just that we did not know how to handle the modelling process in any other way! Limiting process based on smooth models have often been found to have singularities, (for instance in quantum gravity theory in physics), which perhaps shows some limitations for that approach. The fractafolds project's plan  is to use a suitable class of `fractafolds' as a testing ground for a limiting calculus of differential forms  and hopefully the development of corresponding tools.  

\section{Nerves, simplicial complexes and $T_0$-spaces}
As mentioned above, observations may or may not distinguish objects. The structural management of observations can be handled in several different ways.  Although related to each other, these methods do give different information and assume slightly different starting models. The methods either use simplicial complexes or $T_0$-spaces / partially ordered sets.

It is probably a good idea to recall the basic definitions to start with. 

\begin{definition}
A \emph{simplicial complex} $K$ is a set of objects, $K_0$, called \emph{vertices} and a set of finite non-empty subsets of $K_0$, called \emph{simplices}.  The simplices satisfy the condition that if $\sigma \subset K_0$ is a simplex and $\tau \subset \sigma$, $\tau \neq \emptyset$, then $\tau$ is also a simplex.
\end{definition}
\begin{definition}
If $K$ and $L$ are two simplicial complexes a \emph{simplicial mapping} $f: K \to L$ is a function $f_0 : K_0 \to L_0$ in vertex sets that preserves simplices in the sense that if $\sigma\subset K_0$ is a simplex of $K$ then its image, $f(\sigma)\subset L_0$, is a simplex of $L$. 
\end{definition}

The other basic model is that of a finite  $T_0$-space.
\begin{definition}
A topological space $X$ is a $T_0$ space if given distinct points of $ X$, there is an open set of $X$ that contains one but not the other.  
\end{definition}
A $T_0$-space gives rise naturally to a partial order on the set of points of $X$, where $x\leq y$ if for each open set, $U$, of $X$,  $y\in U$ implies $x\in U$ and conversely.

The first method that we will examine is due to Sorkin, \cite{Sorkin:1991} and, roughly speaking, assumes there is a space $X$ being `observed' and that a set of observations correspond to an open cover of the space. This model is `static' as it assumes a given space, but it led later to the causal set approach which is based more on the possible futures of state and is thus dynamic allowing the space being observed to evolve through time more explicitly (see Sorkin's papers and his talk at the workshop).

\textbf{The Sorkin model}, \cite{Sorkin:1991}.

Let $X$ be a space and $\mathcal{F}$, a (locally finite) open cover of $X$.  (The idea of the model is approximately  that open sets correspond to observations, so if $x, x^\prime \in U$, the observation, $U$, `tests positive' on both $x$ and $x^\prime$, so does not distinguish them.)

Using $\mathcal{F}$, define, on the set $X$, an equivalence relation $\sim_\mathcal{F}$, given by\\
\centerline{$x\sim_\mathcal{F} x^\prime$ if and only if, for all $U \in \mathcal{F}$, $x \in U\Leftrightarrow x^\prime \in U$\enspace ,} 
thus two points of $X$ are equivalent if all the observations from $\mathcal{F}$ give the same positive or negative result on them both.  Using $\sim_\mathcal{F}$, we can form a quotient space, $X_\mathcal{F}$.

\textbf{Remarks}

(i) In some ways, this seems silly, since as we do not know $X$, we do not know its topology and so should have little or no knowledge of the quotient topology on $X_\mathcal{F}$. The point is, however, that $X_\mathcal{F}$ \emph{is} something we \emph{do} know.  It encodes the observational data on the mysterious (and perhaps `pointless'), $X$.  The type of simplified model of `observational data', using an idea that `observations behave like open sets', does determine the model, but the type of construction is almost generic.  The space $X_\mathcal{F}$ `organises' the data.

The question of the topology on $X_\mathcal{F}$ initially does look tricky, but quite generally it will be a $T_0$-space and hence should correspond to a partially ordered set in a natural way. The order can be specified without knowing the topology on $X$, merely needing the cover $\mathcal{F}$!  In fact, writing $[x]_\mathcal{F}$ for the equivalence class of $x\in X$, in a very natural way:
\begin{quote}$[x]_\mathcal{F}\leq [x^\prime]_\mathcal{F}$ if and only if, for every open set, $U$ in $\mathcal{F}$, if $x^\prime\in U$, then $x \in U$\enspace.\end{quote}
In fact, in situations such as we are considering, in which the cover is finite, $X_\mathcal{F}$ is a finite $T_0$-space, and each point $[x]_\mathcal{F}$ is in a unique minimal open set, $U_{[x]}$ of $X_\mathcal{F}$, and\\
\centerline{$[x]_\mathcal{F}\leq [x^\prime]_\mathcal{F}$ if and only if  $x \in U_{[x^\prime]}$\enspace.}\\
Of course, this has a nice interpretation in terms of `observations'.  The essential information on $X_\mathcal{F}$ is contained in this partially ordered set and it can be considered to be a spatial representation of the original `space' $X$, relative to the observations considered.

(ii) Later in the second part of this paper, \cite{JGTP:2004b}, we will generalise the Sorkin construction to the abstract setting of formal contexts and Chu spaces and will show that it is a particular case of a well known construction in Chu theory.

\textbf{Nerves}

This partially ordered set is closely related to, but need not be identical with, the partially ordered set of simplices of the nerve of the open cover $\mathcal{F}$. (A detailed examination of the relationship forms part of the second part of this paper, \cite{JGTP:2004b}.)

Recall that given a space $X$, and an open cover, $\mathcal{F}$, the \emph{(\v{C}ech) nerve}, $N(\mathcal{F})$, of $\mathcal{F}$ is defined to be that simplicial complex having the sets of $\mathcal{F}$ as vertices and in which $\{U_0,\ldots, U_n\} \subset \mathcal{F}$ is a $n$-simplex of $N(\mathcal{F})$ if and only if $\bigcap_{i=0}^n U_i \neq \emptyset$.

The construction is a classical one of algebraic topology, cf. \cite{Cech}.

There is an alternative construction, essentially dual to this, and due to Vietoris.  The vertices in this \emph{Vietoris complex} $V(\mathcal{F})$ of $(X,\mathcal{F})$ are the points of $X$ itself and an $(n+1)$-tuple of such points, $\langle x_0, \ldots, x_n\rangle$, is an $n$-simplex if there is a $U \in \mathcal{F}$  that contains them all,
$$\{x_0, \ldots, x_n\} \subseteq U.$$

Dowker showed that these two complexes provide the same information up to homotopy, cf. \cite{dowker}. 

 Dowker's constructions have been rediscovered several times since and one finds similar ideas now being used in Artificial Intelligence, for instance, in Knowledge Representation, cf.,  papers by Giavitto and Valencia, \cite{GV} and Sansonnet and Valencia, \cite{SV}.  In these and elsewhere the basic setup is that which was briefly mentioned in the introduction. There is a set of objects and a set of attributes together with a  relation between them. This situation is well known in Theoretical Computer Science as being a (dyadic) Chu space and in Formal Concept Analysis as a formal context.  (we will tend to emphasise the former theory here for reasons that will become clear later.) We recall the basic definitions so as to have the notion of morphism of Chu spaces  available. (We use \cite{Pratt:Chu:1999,Zhang:ENTCS:2003} as  sources so will combine and adapt the notation used there.)
\begin{definition}
A \emph{(dyadic\footnote{the relation can be thought of as a subset of $P_o\times P_a$ and hence as a map $P_o\times P_a\to \mathbf{2}$, hence the term `dyadic'.  The role of $\mathbf{2} = \{0,1\}$ can be replaced by an arbitrary set $K$ and this leads to a very rich theory.}  or two valued) Chu space} $\mathcal{P}$ is a triple $(P_o, \models_P, P_a)$, where $P_o$ is a set of \emph{objects}, and $P_a$ is a set of \emph{attributes}.  The \emph{satisfaction} relation $\models_P$ is a subset of $P_o\times P_a$.

A \emph{morphism} or \emph{Chu transform} from a Chu space $(P_o, \models_P, P_a)$ to a Chu space $(Q_o, \models_Q, Q_a)$ is a pair of functions $(f_a,f_o)$ with $f_o : P_o\to Q_o$ and $f_a : Q_a \to P_a$ such that for any $x\in P_o $ and $y \in Q_a$,
$$f_o(x)\models_Q y \textrm{ iff } x \models_P f_a(y)\enspace .$$
If $\mathcal{P} = (P_o, \models_P, P_a)$ is a dyadic Chu space, then $\mathcal{P}^\perp = (P_a, \models_P^{op}, P_o)$ is the {dual Chu space} of $\mathcal{P}$. (It just reverses the roles of objects and attributes.)
\end{definition}
 A Chu space when considered  in Formal Concept Analysis (FCA) is usually called a \emph{formal context}.  The level of generality as well as the level of abstraction means that there are many different situations to which they can be applied.  The advantage of the Chu space theory over FCA is that Chu morphisms are an integral part of the theory, whilst the notion of  a morphism of contexts is less well developed.  The rich categorical theory of dyadic,  and more general types of Chu spaces can be found in Pratt's excellent Coimbra lecture notes \cite{Pratt:Chu:1999}. Links between Chu spaces and Formal Concept Analysis are more fully explored in \cite{Hitzler-Zhang:2003,Shen-Zhang:2003,Zhang:ENTCS:2003}. We will see later that in addition to these Chu transforms, there are other useful notions of morphisms, which are particularly suited to the study of nerves.

\begin{definition}
If $\mathcal{P} = (P_o, \models_P, P_a)$ is a formal context or dyadic Chu space, then its (\v{C}ech) nerve is the simplicial complex $N(\mathcal{P})$ with vertex set $P_a$ and where a subset, $\{a_0, \ldots, a_p\}$ of $P_a$ is a $p$-simplex if there is an object $x \in P_o$ satisfying $x\models_P a_i$ for $i = 0, \ldots, p$.

The Vietoris nerve of $\mathcal{P}$ is, by definition, the \v{C}ech nerve of $\mathcal{P}^\perp$.  It will be denoted $V(\mathcal{P})$.  It is worth noting that it has $P_o$ as set of vertices and $\{x_0,\ldots, x_q\}$ is a $q$-simplex if there is an attribute $a$ satisfied by all the $x$s, i.e, for $j = 0, \ldots, q$, we have $x_j\models_P a$.
 \end{definition}

Strangely enough the \v{C}ech and Vietoris nerve constructions do not seem to be functorial on the category of dyadic Chu spaces as no induced simplicial map would seem to exist corresponding to an \emph{arbitrary} morphism of Chu spaces in the obvious way. There is however a generalisation of the induced map for a continuous map of spaces relative to an open cover and this has a nice interpretation in our situation.  Remember that $P_a$ is thought of as a set of attributes and we can think of the relation together with these attributes, as helping us to gain `knowledge ' about the objects.  The set $P_a$ may be large even infinite, and in that case it will be necessary to \emph{sample the attribute set}, thereby choosing a subset $\mathcal{F}$ of $P_a$, and to corestrict the context relation to that subobject getting a formal context $(P_o, \models_P, \mathcal{F})$.

\begin{proposition}
If $\mathfrak{f} = (f_o,f_a) :(P_o, \models_P, P_a) \to (Q_o, \models_Q, Q_a)$ is a morphism of Chu spaces and $\mathcal{F}$ is, this time, a sample of the attributes $Q_a$ of $(Q_o, \models_Q, Q_a)$, then\\
(i)  we have an induced map (which we will also denote by $(f_o,f_a)$), $$(f_o,f_a) :(P_o, \models_P,f_a(\mathcal{F})) \to(Q_o, \models_Q, \mathcal{F})\enspace .$$
(ii) there is an induced simplicial map $$V(\mathfrak{f}) :V(P_o, \models_P,f_a(\mathcal{F})) \to V(Q_o, \models_Q, \mathcal{F})\enspace ,$$
given by 
$$V(\mathfrak{f})\langle p_0, \ldots, p_n\rangle = \langle f_o(p_0), \ldots, f_o(p_n)\rangle \enspace .$$
(iii) Any choice of splitting for the function
$$f_a : \mathcal{F} \to f_a(\mathcal{F}) \hspace{1cm} \mathcal{F}\subseteq Q_a$$
determines a simplicial map
$$N(\mathfrak{f}) : N(P_o, \models_P, f_a(\mathcal{F})) \to N(Q_o, \models_Q, f_a(\mathcal{F}))$$
given by 
$$N(\mathfrak{f})\langle f_a(q_0), \ldots , f_a(q_n)\rangle = \langle q_0,\ldots ,q_n\rangle\enspace .$$
[More exactly, if we choose for each $p \in f_a(\mathcal{F})$, a $q \in \mathcal{F}$ such that $p = f_a(q)$, then $N\mathfrak{f})\langle p\rangle = \langle q \rangle$, and so on for higher dimensional simplicies.]
\end{proposition}
The proof uses just the adjointness relationship\\
\centerline{$f_o(x)\models_Q q$ if and only if $x \models_P f_a(q)\enspace .$}

\begin{corollary}
 If $\mathfrak{f} = (f_o,f_a) : \mathcal{P} \to \mathcal{Q}$ has $f_a$ surjective, then there are induced maps $N(\mathfrak{f}) : N(\mathcal{P})\to N(\mathcal{Q})$ and $V(\mathfrak{f}) : V(\mathcal{P})\to V(\mathcal{Q})$.
\end{corollary} 
\textbf{Remarks}

(i) The question, of course, arises as to what happens if the splitting of $f_a$ is changed.  In this case the two induced maps from $N(\mathcal{P})$ to $N(\mathcal{Q})$ will be homotopic, i.e. each can be `deformed' into the other.

(ii)  The `dual' case when $f$ is surjective can, of course, be handled by duality. In particular if $f$ is the identity on $P_o$,
$$\mathfrak{f} = (id,f_a) : (P_o,\models_P,P_a)\to(P_o,\models_P,Q_a)\enspace ,$$ 
and if we further assume that the Chu spaces are `normal'\footnote{In Pratt's terminology, a Chu space is said to be \emph{normal} if the `attributes' are subsets of the set of objects, so here we are assuming $P_a,Q_a \subseteq \mathcal{P}(P_o)$.}, then $f_a$ will be an inclusion.

The dual morphism 
$$\mathfrak{f}^\perp : \mathcal{Q}^\perp\to\mathcal{P}^\perp$$
has its adjoint part surjective, so $\mathfrak{f}^\perp$ induces a well defined 
$$V(\mathfrak{f}^\perp) : V(\mathcal{Q}^\perp)\to V(\mathcal{P}^\perp) \enspace ,$$
that is,
$$N(\mathfrak{f})^\perp : N(\mathcal{Q})\to N(\mathcal{P})\enspace $$
sending a simplex $\langle q_0, \ldots, q_n\rangle \in N(\mathcal{Q})$ to $\langle f_a(q_0), \ldots, f_a(q_n)\rangle \in N(\mathcal{P})$.  Similarly a choice of splitting for $f_o$ gives a simplicial map
$$V(\mathfrak{f}^\perp) : V(\mathcal{Q}) \to V(\mathcal{P})$$
with different splittings giving homotopic maps.

(iii) There is another situation that classically leads to an induced map on the nerves, namely, refinement.  This does not seem to be subsummed under the Chu transform model and we will return to it later.

\pagebreak

\textbf{A Critique}

Each of these methods for exploring the interrelations between objects and attributes has its advantages and its disadvantages.  If the carrier set $X$ underlying the Chu space (that is, the object set of the context) is essentially unknown (perhaps even unknowable), then the Sorkin construction and the Vietoris construction become problematic unless adapted, (see later, in the second paper). The \v{C}ech nerve would make sense even in a `pointless' context, but the points or objects in the space serve to `rigidify' the changes when the cover is refined. We give this initially just in the classical context of a space with two open covers:\begin{quote}if $\mathcal{F}^\prime$ is finer that $\mathcal{F}$, then, by definition, for any $U \in \mathcal{F}^\prime$, there is at least one $V \in \mathcal{F}$ with $U\subseteq V$.\end{quote}
 
As we will see later there are at least two different and useful types of refinement in this context.  When we need to we will refer to the above as \emph{ \v{C}ech refinement}.

A problem occurs with refinement and  the \v{C}ech nerve construction if, for some $U \in \mathcal{F}$, there are several such 
$V$, then there will be a simplicial map
$$N(\mathcal{F}^\prime)\to N(\mathcal{F})$$ obtained by making a choice of one such $V$ for each $U$, but  different choices give different maps so the induced map is \emph{not} determined by the refinement process.
  For the Sorkin and Vietoris models,  on the other hand, there is an obvious natural map
$$X_{\mathcal{F}^\prime}\to X_{\mathcal{F}}$$
resp.
$$V({\mathcal{F}^\prime})\to V({\mathcal{F}})$$
induced by the refinement relation, without need of a definite choice of `refinement map', however this depends on having `points'.

Computationally with data in the form of a point cloud, this problem is transmuted to another.  An open cover is derived from a sample of the data cloud.  Algorithms are used (perhaps nearest neighbour, or similar and perhaps with clustering) to produce a simplicial complex / triangulation that approximates the space and if an open cover is produced it is of that complex, not of the original space.  The points that witness to, say, the non-emptiness of an intersection, are estimates of points `in the space'.  It is as if $\langle U_0, \ldots, U_n\rangle$ is an $n$-simplex if $\bigcap U_i$ is \emph{observed} to be non-empty.  Refinement here may typically mean taking a larger sample including the old one.  This may result in the detection of holes, but as with the \v{C}ech  nerve, problems arise in the refinement process. (For an overview of some of the problems in this area, see the survey article by de Floriani, Magillo and Puppo, \cite{FMP}.)

Variants of both the \v{C}ech nerve and the Vietoris complex has also been used  in this context and in the theoretical development of feature identification algorithms, see \cite{carlsson-carlsson-daSilva}.  

Suppose that $X$ is a metric space with metric $d$.  For any finite subset $S$ of $X$ and any $\epsilon > 0$, define the \emph{$\epsilon$-Rips complex of $S$} to be the abstract simplicial complex whose vertex set is $S$ and where a subset $\{s_0, \ldots, s_k\}$ is a simplex if and only if $d(s_i,s_j) \leq \epsilon$ for all $i$, $j$ with $0\leq i,j\leq k$.  

Dually, and in the same situation, consider $\mathcal{U}(S)_\epsilon$ to be the collection of open balls in $X$ of radius $\epsilon$ and with centres in $S$. The nerve of this family gives a second complex, $C_\epsilon(S)$, the \v{C}ech complex of the sample $S$ at resolution $\epsilon$.

The intuition is that $X$ is the `feature space', typically thought of as a subset of $\mathbb{R}^n$ for a high value of $n$, giving possible position coordinates plus some extra ones specifying attributes, and $S$ is a sample from the point cloud data taking values in $X$. We therefore think of $S$ as a subset of $X$.   The complex approximated the space being `observed'.  The computational advantage of the Rips complex is that the test $d(s_i,s_j) \leq \epsilon$,  lets one identify the 1-simplices (edges) which then determine the rest of the complex.  A lot of work has gone into the question of the density of sampling from the point cloud. Too low a density  of sampling and there will be small features that are missed.  Features near which, say, the curvature is changing rapidly may need denser sampling, yet too high a sampling level not only leads to slower computation, but can also include distortion of the basic geometry of the spatial representation even to the extent of changing the dimension. (In the Vietoris complex, the dimension of the realisation is often unbounded.  The dimension  of the Rips complex will depend on the local size of the sample used etc.) It is however possible to attempt useful geometric feature identification, to use isosurface algorithms and to find a good continuous representation of the original object in many cases, although the initial data was discrete and obtained from sampling data that originated, say, from a medical CT or MRI scanner.

If we examine these refinement problems through the abstract perspective of Chu spaces, we arrive at a definition of `\v{C}ech' refinement for Chu spaces, and we will look at this in detail shortly.

In more generality, any continuous $f : X \to Y$ between spaces, together with open covers $\mathcal{U}$ of $X$ and $\mathcal{V}$ of $Y$, will induce a simplicial map on nerves if $\mathcal{U}$ is finer than $f^{-1}(\mathcal{V})$. We abstract this via a notion of a refinement relation relative to a carrier function (on objects), the absolute version being the case when $f$ is the identity function.

\textbf{Definitions}

Given $\mathcal{P} = (P_o,\models_P,P_a)$ and $\mathcal{Q} = (Q_o,\models_Q,Q_a)$ and a function $f = f_o: P_o\to Q_o$, which we wil call a \emph{carrier function}, a (\emph{\v{C}ech}) \emph{refinement relation}    relative to $f$, from $\mathcal{P}$ to $\mathcal{Q}$ is a relation
$$\rightarrow_f\subseteq P_a\times Q_a $$
such that
\begin{center}
if $x\models_Pp$ and $p\rightarrow_f q$, then $f(x)\models_Q q$\enspace .
\end{center}

A \emph{(\v{C}ech) refinement map} for a given carrier function $f$) is a function $\rho :P_o\to Q_o$ such that \\
\centerline{for all $p\in P_o$, $p\rightarrow_f \rho(p)$\enspace ,}
i.e. 
\centerline{if $x\models_P  p$ then $f(x)\models_Q \rho(p)$\enspace .}

\textbf{Remark}

There would seem to be a maximal such relation given by:\\
\centerline{$p\rightarrow_f q$ if and only if $\{ x | x\models_P p\} \subseteq\{x | f(x)\models_Q q\},$}
however for the moment we will restrict our attention to the general form.

The relationship with Chu transforms is given by:
\begin{proposition}
(i) If $\mathfrak{f} = (f_o,f_a) : (P_o,\models_P,P_a) \to (Q_o,\models_Q,Q_a)$ is a Chu transform,
 $$ p\rightarrow_f q \textrm{ if and only if } p = f_a(q)$$
is a \v{C}ech refinement relation relative to the carrier function, $f_o$.\\
(ii) If $\mathfrak{f} = (f_o,f_a)$ and $f_a$ is surjective, any splitting $\rho :P_a\to Q_a$ of $f_a$  is a \v{C}ech refinement  map for the carrier function $f$.
\end{proposition}

It is important to remember that the definition of $\rightarrow_f$ does not assume that $\rightarrow_f$  is non-empty, nor `total' in the sense that for each $p\in P_a$, there is a $q \in Q_a$ such that $p\rightarrow_f q$, but  that frequently in the situations we will consider `totality' is a natural condition to assume.

We can extend any such relation $\rightarrow_f$ to one
$$\rightarrow_f\subseteq Fin( P_a)\times Fin(Q_a) $$
by defining, for $X \in Fin(P_a)$ and $Y \in Fin(Q_a)$,
\begin{quote}$X\rightarrow_f Y$ if and only if, for all $p \in X$, there is a $q\in Y$, with $p\rightarrow_f q$\enspace . \end{quote}
Here  $Fin(A)$ denotes the set of finite subsets of $A$.

There would seem to be a link here with the approximable morphisms of formal contexts studied by Shen and Zhang, \cite{Shen-Zhang:2003} and Hitzler and Zhang, \cite{Hitzler-Zhang:2003}.  The following certainly holds:
\begin{proposition}
For any carrier function $f : P_o\to Q_o$,
$$\emptyset \rightarrow_f \emptyset$$
and\\
\centerline{ if $X\rightarrow_f Y_1$ and $X\rightarrow_f Y_2$, then $X\rightarrow_f Y_1\cup Y_2$\enspace .}
\end{proposition}
The proof is easy.  The final condition for approximable morphism does not hold in general, but a variant does, namely,

\centerline{ if $X_1\subset X_2$, $X_2\rightarrow_f Y_1$ and $Y_1\subseteq Y_2$ then $X_1\rightarrow_f Y_2$\enspace .}

\section{Back to Fractafolds.}

At present no definition of a fractafold as such exists. There are examples and conjectured cases that we would hope to include.  The `bootstrapping' process is to develop the tools we believe might analyse these spaces, and see what spaces can be analysed in the process, hopefully inching towards a workable definition. The Menger cube and the Menger manifolds are prime candidates as are the solenoids and more conjecturally some of the well known fractal spaces such as the Lorentz attractor and the R\"{o}ssler band. (The latter is defined by certain linked very simple differetnial equations, namely 
\begin{eqnarray*}\frac{dx}{dt} &=& -(y + z)\\
\frac{dy}{dt} &=& x + ay\\
\frac{dz}{dt}& = &b + xz -cz\enspace ,
\end{eqnarray*}
where $a$, $b$ and $c$ are parameters, for some values of these parametres, nothing exceptional happens,
 but the graphics of the attractor for other values seem to show a M\"{o}bius band-like space having the interval replaced by a Cantor set.\footnote{The values $a=$ 0.15, $b=$ 0.2, $c=$ 10.0 will do and such images and some interesting demos can be  found on the website http://bill.srnr.arizona.edu/demos/rossler/rossler2.html or by a simple search on the web})  

In general, such spaces are of interest as they occur in modelling situations representing the limiting behaviour of some dynamical system, but from the point of view of spatial representation they also provide a challenge that will test to the limit the mathematical tools available for the extraction of geometric (and not just topological) information from the spatial specification.  For instance, these spaces may curve and twist, but there seems no machinery at present that will measure those characteristics, since traditional differential geometry needs smoothness. 

As a starting point,  we do have some intuition about these `fractafolds'.  As was said earlier, they should have global structure that is fairly regular, but locally may be specified by some IFS or similar.  If a space is not globally fairly `homogeneous', then ideally the tools should idntify the regions of `irregularity', but in the first instance, we do not want to `bite off more than we can chew'. This means that there should be both global and local structure that is sufficiently regular but of somewhat different natures. (One can construct artificial examples by taking a Menger cube  and a smooth curve or surface and constructing their product. The dyadic solenoid is fibred over the circle with fibre a Cantor set, and so on.)

For such `fractafolds', the above discussion of nerves leads to a natural desire to use the local regularity to control the  open covers being used as much as possible.  As a `toy' example, the Cantor set is typical.  There are obvious open covers that one can use.  Each stage of the construction of a `middle third' Cantor set yields a polyhedron and the self similarity / iterated function scheme yields computational methods of refinement that can construct the next stage of refinement from the current one. 

\textbf{Remark}

 Note that here  polyhedra give  covers that reflects their combinatorial structure and that, on restriction to the original limiting space, give covers that have nerves whose realisations are essentially the original approximating polyhedra.
These covers, the \emph{star open covers}, of a polyhedron $K$ are obtained from a triangulation by taking, for each vertex, $v$, the union $U_v$ of all open simplices having $v$ as one of their vertices.  The cover is then $\{U_v : v \textrm{ a vertex of the triangulation of } K\}$.  The nerve of such a cover is then essentially the same as the  abstract simplicial complex underlying the triangulation.

For the Cantor set polyhedral approximation 
$$[0,\frac{1}{3}]\cup [\frac{2}{3},1]\enspace ,$$
the cover will be 
$\{[0,\frac{1}{3}),(0,\frac{1}{3}],[\frac{2}{3},1),(\frac{2}{3},1]\}$, and so on.  The nerve in this example is obvious, whilst the poset model is 
$$\xymatrix{\bullet &\bullet \ar[l]\ar[r]&\bullet & \quad &\bullet &\bullet\ar[r]\ar[l] &\bullet }$$
or its dual, depending on the conventions being used.

(In the second part of this paper, we will return to this example  in more detail.)

\medskip

\section{`In the limit \ldots '}

The intuition behind these discrete representations is that, as the `mesh' of the cover or triangulation tends to zero, the space $X$ being modelled emerges as the limit.  In models of physical systems, this is philosophically and physically problematic, but is none the less useful.  The limit space should perhaps be thought of as providing insight into the system of approximating `data objects'.  The `points' of the `space' are ideal entities.  This viewpoint is essential in computational situations, where the `points' in the limit represent infinite processes, hence, once again, are idealised models of the objects of interest.

`Taking a limit' can only occur within a mathematical context, in fact, within a category  and the answer may,  and often will, be dependent on which category you take that limit in.  Perhaps from this point of view, a `fractafold' will be a particularly well structured approximating system of combinatorially defined spaces - but a `manifold' will also be represented by a different, but quite similar, structured approximating system of combinatorially defined spaces!

One common example of a limit space, much used when talking about computational issues is, of course, the Cantor set in its various guises.  The basic Menger cubes are $n$-dimensional analogues of the Cantor set.  As you refine the obvious open covers, ever more holes or handles are revealed at the finer scale.  A Menger manifold is locally like a Menger cube of fixed dimension, so you expect initially that refinements of open covers of such objects likewise will reveal more holes or handles, but that it is not clear that any other complications arise.  It is even known that Menger $n$-manifolds have a fibre structure that makes them look, for many purposes, locally like a product of a polyhedron with a Menger cube (see the book by Chigogidze, \cite{Chig}).  Menger manifolds behave as if they were the finite dimensional analogues of Hilbert cube manifolds, and those have a very well behaved topology. All this would suggest that Menger manifolds should be easy to study and would not give us many problems.  The reality is that other phenomena can occur within the candidate spaces that are slightly unexpected from the viewpoint of observations, open coverings, etc.

Some of these potential peculiarities of `fractafolds' can be illustrated by the class of spaces known as solenoids.  These seem to be very closely related to the Menger manifolds, but they show clearly how our intuitions may need a bit of `refining'.

\textbf{Solenoids}

The solenoids have similar structure, and are very simple to construct.  They arise as the attractors of dynamical systems and are closely related to the Lorenz attractor and the Rossler band. 
In general, a solenoid, $M_\infty$, is an inverse limit space of a sequence of closed manifolds $M_i$ with `bonding' or `structure' maps
$$p_i : M_{i+1} \to M_i$$
for $i \in \mathbb{N}$, which are covering maps, in the topological sense, such that any composite
$$p_{i+k}\circ \ldots \circ p_i : M_{i+k+1} \to M_i$$
is a regular covering map.  Solenoids are homogeneous spaces, so if $x, y \in M_\infty$, there is some homeomorphism $h : M_\infty \to M_\infty$ such that $h(x) = y$.  (Locally they are similar everywhere, so this looks after the regularity issue here.)

We will not look at this general case, but restrict to one example, the \emph{dyadic solenoid}, $DS$.  To construct this, we take each $M_i$ to be the unit circle $S^1$, and each $p_i$ to be a degree two covering map, (double covering).  (To be explicit, represent $S^1$ as the space of unit modulus complex numbers and $p_i(z) = z^2$, the map wrapping $S^1$ around itself twice.)  The points of DS can be given an explicit description as sequences $(z_i)$ of unit moduli complex numbers with $z_{i+1}^2 = z_i$ for each $i \in \mathbb{N}$.

There is a well known representation of $DS$ as an intersection of solid tori in $\mathbb{R}^3$.  Take a solid torus in $\mathbb{R}^3$ and within it put a second solid torus that wraps around itself twice.

There is, of course, a homeomorphism from the big torus to the small one, that maps the initial small one to an even smaller one that now wraps itself four times around the original central hole.  Repeat \emph{ad infinitum}! The intersection of all these ever smaller solid tori is the dyadic solenoid.  Points to note include:\\
(i) At any observational resolution, the space, $DS$, will seem to be a solid torus, but the effect of refining the open cover/observations will be, not just, to see finer detail, but also to observe the double covering or its iterates.\\
(ii) There is a projection map $p_\infty : DS \to S^1$, mapping the sequence $(z_i)$ to $z_0$, and this is a principal bundle with fibre a Cantor set, $CS$.  Thus locally this map looks like a product from `$CS\times$ interval' to the interval, yet the way the product twists as one goes around the circle produces the strange features of the solenoid.
( A detailed analysis of this twisting would bring in the profinite group of automorphisms of $CS$.)

At any scale one has a manifold-like space, but at the limit of computation, it is a highly singular space.  That space is `locally self similar', even homogeneous.

\textbf{Remark}

It is worth repeating that although artificially generated, the dyadic solenoid shows many characteristics of strange attractors that \emph{are} encountered in mathematical models for physical phenomena.  Those models use smooth tools, being based on differential equations, but to understand the geometry of the model, we have to take seriously the `continuous vs. discrete' transition involved in a) scientific computation and modelling, and  b) scientific visualisation and scientific observation

\section{Geometry?}

Static representation of spaces, or alternatively representation by static spaces, is not the end of the problem.  A dynamic approach to the modelling/representation problem is needed, since actual spaces change in time. Geometric structures such as vector fields, differential equations or their analogues, and dynamical systems on a space allow the modelling of change \emph{on} a space. There is also change between spatial representations at different observational scales and, for the finer analysis of geometric features, some analysis of that change will be needed.

The approach to such geometric structures is usually modelled on the differential paradigm, but the usual tools of differential geometry such as curvature, torsion, etc., depend on smooth structure. The approximating spaces naturally arising from the discretisation / measurement / computational framework are typically simplicial complexes.  One key tool of smooth geometry is the \emph{differential form} however, and there are several possible ways in which these can be extended from the smooth to the simplicial / discrete context.

\textbf{Differential Forms.}

The classical situation is based on a smooth manifold, $M$, with tangent bundle $TM$. (We use Janich, \cite{janich} as a basic reference.)  For convenience we recall some elementary definitions from multilinear algebra and differential geometry.  

For $V$ a real vector space, an alternating $k$-form $\omega$ on $V$ is a map $$\omega :\underbrace{V\times \ldots \times V}_k \to \mathbb{R}$$
that is linear in each variable and in addition has the property that $\omega(v_1, \ldots, v_k)= 0$ if the vectors $v_1, \ldots, v_k\in V$ are linearly dependent.  We will denote the vector space of alternating $k$-forms on $V$ by $Alt^kV$. 

A form of degree $k$ on $M$ is a function that assigns to every $p \in M$, an alternating $k$-form $\omega_p \in Alt^kTM_p$ on the tangent space to $M$ at $p$. (To ensure that the assignment makes sense globally, we actually form a vector bundle $Alt^kTM$ in the obvious way by applying the $Alt^k$ to each fibre, and then  $\omega$ is a continuous `section' of that vector bundle.) 

If we have a given chart at $p$, then the coordinate directions give a basis for $TM_p$.  The elements of this basis will be written $ \partial_\mu = \partial/\partial x^\mu$. The space of 1-forms has a dual basis with elements $dx^\mu$.  Any $k$-form $\omega$ can locally be expanded in terms of symbols $dx^{\mu_1}\wedge  \ldots\wedge  dx^{\mu_k}$ with coefficients $\omega_{\mu_1, \ldots,\mu_k}: = \omega(\partial_{\mu_1}, \ldots, \partial_{\mu_k})$.  If these functions are differential with respect to the given coordinate chart, then $\omega$ is called a differential $k$-form.  The space of these is denoted $\Omega^kM$.

\textbf{Examples and explanation of notation.} 

If $f: M \to \mathbb{R}$ is a differentiable function, then $df : TM \to T\mathbb{R}, df(X) = X(f) $ is a differential 1-form, once we identify $ T\mathbb{R}$ with $\mathbb{R}$ in the canonical way. Here $X$ is a vector field on $M$ and is thus a section of the tangent bundle $TM$. It is this that gives the exact meaning of the notation $dx^\mu$ used earlier.  The wedge or exterior product of vector spaces is a usual construction of multilinear algebra.  Here it allows us to write any $\omega \in \Omega^kM$ locally as a sum 
$$\omega = \sum \omega_{\mu_1, \ldots,\mu_k}dx^{\mu_1}\wedge  \ldots\wedge  dx^{\mu_k}\enspace .$$
As an elementary example consider $M \subset \mathbb{R}^3$ an open set and therefore a 3-manifold. The usual volume element $dV$ of elementary  integration and vector calculus is given by $dx^1\wedge dx^2\wedge dx^3$.  Similarly the usual vector calculus operations,  \emph{div}, \emph{grad} and \emph{curl}, have easy descriptions in terms of differential forms, cf. \cite{janich}, p.172.
If $M$ is an $n$-dimensional manifold then $TM_p$ is an $n$-dimensional vector space, and so $\Omega^kM = 0$ if $k >n$.

If $\omega \in \Omega^kM$,   we can form its Cartan derivative $d\omega \in \Omega^{k+1}M$, which is locally given by
$$d\omega = \sum d\omega_{\mu_1, \ldots,\mu_k}\wedge dx^{\mu_1}\wedge  \ldots\wedge  dx^{\mu_k}\enspace .$$
It is moderately easy to check that $dd\omega = 0$, so we get a chain complex
$$0\to \Omega^0M\stackrel{d}{\to}\Omega^1M\stackrel{d}{\to}\ldots \stackrel{d}{\to}\Omega^{n-1}M\stackrel{d}{\to}\Omega^nM\stackrel{d}{\to}0\enspace  .$$
This is the \emph{de Rham complex} of $M$.  One can use the exterior product to multiply forms together and the differential is compatible with this multiplication in a fairly simple way, (see later), so it is a \emph{graded differential algebra} or \emph{dga}, see \cite{janich,halperin,lehmann}.  The quotients
$$H^r_{dR}(M) = \frac{Ker(d : \Omega^rM\to \Omega^{r+1}M)}{Im(d : \Omega^{r-1}M\to \Omega^rM)}$$
form a cohomology theory and the famous de Rham theorem says that these geometrically defined groups are isomorphic to the topologically defined singular cohomology groups with real coefficients.  This, loosely speaking, tells one in what way the \emph{geometric} differential forms react to, or reflect,  the overall \emph{topological} structure of the space, $M$.

One plan of action for the study of `differential forms' on fractafolds and thus for a geometry of information change, is to look at candidates for theories of differential forms on non-smooth spaces and, initially, to apply the test: is there a reasonable substitute for the de Rham theorem in that theory?
\medskip

\textbf{Technical remark}

It is often useful to replace the simplicial complex / polyhedral models for approximations by ones using simplicial sets.  The detailed theory of simplicial sets need not concern us here, but if one orders the vertices of a simplicial complex in such a way that in each simplex the vertices are totally ordered, then one can encode constructions such as products of objects using the simple tool of degenerate simplices.  For instance, the simplicial complex corresponding to the interval has two vertices 0, and 1 and one 1-simplex $\{0,1\}$.  If we order the vertices $0 < 1$, and then allow degenerate simplices such as $(0,1,1)$ or $(0,0,1,1,1)$ by repeating a vertex label to get the simplicial set $\Delta[1]$, we can represent the simplicial structure of a square
$$\xymatrix{(0,1)\ar[r]&(1,1)\\
(0,0)\ar[r]\ar[u]\ar[ur]^{(a)}_{(b)}&(1,0)\ar[u]} \quad \textrm{ where } (a)= \left(\begin{array}{rrr}0&0&1\\ 0&1&1\end{array}
\right) = \big((0,0,1),(0,1,1),\big) \textrm{ etc.,}$$
by the product of two copies of $\Delta[1]$, where, for simplicial sets $K$, $L$, the product simplicial set $K\times L$ has its collection of $q$-simplices given by $(K\times L)_q = K_q\times L_q$ as a product of sets.

Another useful difference between simplicial sets and simplicial complexes is that simplices now have definite labelled faces.  If $\sigma = \{a_0,a_1,a_2,a_3\}$ is a 3-simplex in a simplicial complex, then we know its 2-dimensional faces are all 2-element subsets of $\sigma$, but describing them in more detail is tricky.  If we had that $\sigma$ was a 3-simplex $( a_0,a_1,a_2,a_3) $ in a corresponding simplicial set, so that the vertices were in the given order, then $\sigma$ has faces $d_0\sigma = (a_1,a_2,a_3)$, $d_1\sigma = (a_0,a_2,a_3)$, etc., where the $i^{th}$ face $d_i\sigma $ leaves out the $i^{th}$ entry in the list of vertices.  (Simplicial sets have been explicitly used in computer graphics in an attempt to combine the advantages of both cubical and simplicial grids using products of them to get a combinatorial grid model including prisms.)

\medskip

The theme of `continuous vs. discrete' repeats in the classification of the models for the analogues of the  de Rham complex that are applicable to simplicial complexes. (We will not be exhaustive in our listing of these and will choose three or four, looking at one in a lot more detail in the second part of this paper.)

\textbf{`Continuous' geometric models} (Whitney-Thom-Sullivan)

Let $\delta^n$ be the standard Euclidean $n$-simplex
$$\Delta^n = \{ \underline{t} \in \mathbb{R}^{n+1} ~|~ \sum t_i = 1, \textrm{ all }t_i \geq 0\},$$
 and let $A_{dR}(\Delta^n) $ be the algebra of differential forms defined on neighbourhoods of $\Delta^n$ in the hyperplane given by $\sum t_i = 1$. If $K$ is any simplicial set, we can take a copy of $A_{dR}(\Delta^n) $ for each $n$-simplex, $\sigma\in K_n$ and repeating for each $n\geq 0$, `glue them together' along shared faces to get the algebra
$$A_{dR}(K) = Hom(K,A_{dR}(\Delta^\bullet)).$$
There is an important point to note here.  There are obvious face and degeneracy maps defined between the various spaces $\Delta^n$.  For instance, there is the inclusion of an $n-1$-face with, say, the $i^{th}$ coordinate $t_i = 0$.  Correspondingly, by restriction, you get  maps of algebras, e.g., $d_i : A_{dR}(\Delta^n)\to A_{dR}(\Delta^{n-1})$. (Note the direction of the arrow.) This gives $A_{dR}(\Delta^\bullet)$ two structures.  It is a simplicial object in the category of differential graded algebras. In the `Hom' above we use the simplicial structure, and this leaves this `hom-set' to inherit the differential graded algebra structure. This means that defining models for de Rham type complexes we really only need to define them on simplices and then we can use the same trick to extend them to all simplicial sets.  This is a categorical  `density' result. Defining some functor / construction on a dense full subclass is sufficient to define it everywhere.  (A detailed treatment of these ideas is given in  \cite{halperin,lehmann},  or in several more recent texts.)
\medskip

\textbf{Theorem} (Simplicial de Rham, cf. \cite{lehmann}, p.41)

\emph{The associated homology of $A_{dR}(K)$ is isomorphic to the cohomology of $K$ with real coefficients.} 

\medskip

\textbf{`Sheaf theoretic' geometric  models,} cf.  Mostow, \cite{mostow}

Various sheaf theoretic extensions have been suggested.  That which we will describe here is due to Mostow.  He bases his construction on a notion of differentiable space that may be useful in its own right, so here it is:

\textbf{Definition:}

A \emph{differentiable space} is a topological space $X$ together with for each open set $U$ in $X$, a collection, denoted $C^\infty(U)$ of continuous real-valued functions of $U$, satisfying the following `closure' conditions:
\begin{enumerate}[(i)]
\item The assignment $U \rightsquigarrow C^\infty(U)$ defines a sheaf on $X$, which will be denoted $\mathbf{C}^\infty(X)$;
\item For any $n$, if $f_1, \cdots, f_n\in C^\infty(U)$ and $g \in C^\infty(\mathbb{R}^n)$ (with the usual meaning), then $g(f_1, \cdots, f_n) \in C^\infty(U)$.
\end{enumerate}

A basic way to define a differentiable space structure is the following.  Let $X$ be a topological space and let $\{f_a : U_a \to M_a\}$ be a collection of continuous functions from open subsets $U_a$ covering $X$ to manifolds $M_a$.  A function $f : U \to \mathbb{R}$ ($U$ open in $X$) will be said to be \emph{locally a smooth function of finitely many of the $f_a$} if for each $x\in U$, there exist a neighbourhood $W$ of $x$ in $U$, a finite set of indices $a_1, \ldots, a_n$ and a smooth map $g : V\to \mathbb{R}$ (where $V$ is open in $M_{a_1} \times \ldots \times M_{a_n}$) such that for each $i = 1,\ldots, n$,
\begin{enumerate}
\item $f_a$ is defined on all of $W$ (i.e. $U_a\supset W$),
\item $f|W = g\circ (f_{a_1}, \ldots, f_{a_n})$.
\end{enumerate}
Let $C^\infty(U)$ be the set of all such $f$.  Then $\{X,\{C^\infty(U)\}\}$ defines a differentiable space structure on $X$.

\medskip

\textbf{Examples}

1. A smooth manifold $M$ with its usual collection of (locally defined) smooth functions is a differentiable space.

2. A topological space $X$ becomes a differentiable space in a trivial way if we define every continuous function on $X$ to be smooth, $C^\infty(U) = C(U)$.
  
3. A simplicial complex $X$  becomes a differentiable space if every function on $X$, which is locally  a smooth function of finitely many barycentric coordinates is called smooth, so we could use our approximating simplicial complexes, give them differentiable spaces structures and attempt to `pass to the limit'.  (In fact we will not explore that line in detail here, but it would be an interesting  one to pursue.)

\medskip

A \emph{morphism} or \emph{smooth map} of differentiable spaces
is a continuous map which pulls back smooth functions to smooth functions.  That is $h : X \to Y$ is smooth if \\
(i) $h$ is continuous,\\
(ii) for all open $U\subseteq Y$ and $f \in C^\infty(U)$, $f\circ h\in C^\infty(h^{-1}U)$.

\medskip

If $M$ and $N$ are smooth manifolds with their usual differentiable space structure, then $f : M \to N$ is a morphism of differential spaces if and only if it is a smooth map in the usual sense.

\textbf{The de Rham complex of a differentiable space}

Mostow, \cite{mostow}, does not define a notion of tangent vector for differential spaces, rather he defines differential forms.  These are abstract symbols $$\sum f_{\alpha_0}df_{\alpha_1}\wedge \ldots \wedge df_{\alpha_k}\enspace ,$$ where each $f_{\alpha_i}\in C^\infty(U)$, $U\subset X$.  This requires one or two subsidiary notions to make this more precise .

Let $X$ be a differentiable space.  Then a \emph{plot} of $X$ is a smooth map $\phi :E \to X$, where $E$ is an open subspace of $\mathbb{R}^n$ for some (finite) $n$. 

Let $U$ be a differentiable space, and let $f_{ij}\in C^\infty(U)$, $i = 1, \ldots, p$; $j = 0, \ldots, q$. Let $\eta$ denote the symbol $\sum_{i=1}^pf_{i0}df_{i1}\wedge \ldots \wedge df_{iq}$, and let $\phi : E \to U$ be a plot.  Then $\phi^*\eta$ will denote the differential form
$$\sum_{i=1}^p(f_{i0}\circ \phi)d(f_{i1}\circ \phi)\wedge \ldots \wedge d(f_{iq}\circ \phi)\in A^q(E)\enspace .$$
Let $B^q(U)$ be the real vector space of symbols of this form ($p$ arbitrary) modulo the equivalence relation:\\
\begin{quote}$\eta_1\sim\eta_2$ if and only if $\phi^*\eta_1 = \phi^*\eta_2$ for all plots $\phi : E \to U$\enspace .\end{quote}
If $X$ is a differentiable space, then the rule $U\mapsto B^q(U)$, $U$ open in $X$, is a presheaf on $X$. We set $\mathbf{A}^q(X)$ to be the sheaf generated by this presheaf and let $A^q(U) = \Gamma(\mathbf{A}^q(X)|U)$. (As usual $\Gamma$ indicates the space of global sections of a sheaf.) The passage from $B^q$ to $A^q$ means that the latter contains not only the finite sums of forms that are in $B^q(X)$, but also locally finite ones, i.e. you may have an infinite sum but only finitely many terms are non-zero at any one place. The natural map from $B^q(U)$ to $A^q(U)$ \emph{is} an inclusion. 

The commutative dga $A^*(X) = \bigoplus_q A^q(X)$ might be called the (Mostow-)de Rham complex of the differentiable space $X$.  The de Rham cohomology of $X$, $H^*_{dR}(X)$ is the cohomology of the underlying cochain complex.
Again a version of the de Rham theorem holds here.

\medskip

\textbf{Discrete Models}

Passing to discrete models as against continuous ones, there is a halfway stage using polynomial models for the simplices that use the commutative rings, $\mathbb{R}[t_0, \ldots, t_n]/(\sum t_i -1)$, but these will not be considered here.  (A treatment  of such polynomial based forms is given in Karoubi's work, \cite{kar1,kar2}, which contains a discussion of a de Rham type theorem.)

A more purely discrete model is based on the incidence algebra of a partially ordered set.  There are many different versions of this, most of which are closely related to each other. We will here introduce a version due to Zapatrin,  \cite{zapatrin}, and will explore the internal structure of this model in a quite a lot of detail in the second part of this paper.

\medskip

\textbf{The combinatorial incidence algebra}

There are various variants of the construction of  differential forms algebraically but most of them use as a first step the incidence algebra of a graph or of a poset.  This construction is well known from combinatorics and there is an extensive theory in the literature.  Here we can do little more than to give the definition.  We assume given a fixed field $k$, which will usually be $\mathbb{R}$ or $\mathbb{C}$ in our context.

Suppose $(K, \leq)$ is a partially ordered set and form the vector space, $\Omega(K)$ spanned by all pairs $P\leq Q$ in $K$.  If we take $e_{P\leq Q}$ as the basis element corresponding to $P\leq Q$, we get a multiplication on $\Omega(K)$ 
by extending the rule
$$e_{P\leq Q}.e_{R\leq S} = \left\{\begin{array}{cl}
0 & \textrm{ if } Q\neq R\\
e_{P\leq S} & \textrm{ if } Q = S
, \end{array}\right.$$ using linearity.
We will normally restrict attention to finite posets and for such it is clear that multiplicatively everything in $\Omega(K)$ is built up from the `irreducible' pairs $P\leq Q$ for which if $P\leq R\leq Q$,  then either $R=P$ or $R = Q$.
This algebra, $\Omega(K)$, called the \emph{incidence algebra} of $K$, is a $k$-algebra, but is usually non-commutative.

There are other approaches to this construction, some of which are useful in our context.  Any poset $K$ can be considered as a small category in a well known way and given any small category $\mathcal{I}$, one can form the free $k$-linear category $k(\mathcal{I})$ on $\mathcal{I}$.  This is a ring with several objects in the sense of Mitchell, \cite{mitchell}, and again there is a well known construction that takes the associative ring or algebra consisting of the arrows of $k(\mathcal{I})$ with the multiplication extending the partial multiplication / composition of $k(\mathcal{I})$
by defining $a.b = 0$ if the composite $a\circ b$ is not defined in $k(\mathcal{I})$.  This, of course, just gives the incidence algebra back if $\mathcal{I}$ is the small category corresponding to a poset.

There is another slant  on incidence algebras via directed graphs.  Given any directed graph, $\Gamma$, we can form 
the free category of paths on $\Gamma$, which we will denote by $\mathcal{C}(\Gamma)$  Then the obvious thing to do is to form $k\mathcal{C}(\Gamma)$ and from there the corresponding incidence algebra.  It is worth noting that if we have a partially ordered set $(K,\leq)$ then the Hasse diagram, $\Gamma_K$, of $(K,\leq)$ is a directed graph and we can apply the above construction to it.  There is a quotient functor from $\mathcal{C}(\Gamma_K)$ to $K$ itself, and hence an epimorphism from the incidence algebra of $\Gamma_K$ to that of $K$.

\medskip

From any poset, one can form its nerve, which is the simplicial set with strings of elements $P_0\leq \ldots \leq P_n$ as its $n$-simplices. Going the other way around, from a simplicial complex $K$, one can form the poset of its faces.  The nerve of this poset corresponds to the barycentric subdivision of the original $K$.

For example,  the unit interval / 1-simplex complex $\Delta[1]$ has poset
$$
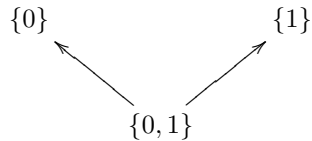
\xymatrix{\{0\}&&\{1\}\\&\{0,1\}\ar[ul]\ar[ur]&
}$$
and this has nerve
$$\xymatrix{\bullet&\bullet\ar[r]\ar[l]&\bullet}$$
which is the barycentric subdivision of the original `interval'.

\medskip

\textbf{From incidence algebras towards differential graded algebras.} 

(Although much of this section is well known, we have included it as certain aspects may not be known from the viewpoint we require.)

As a directed graph is just simplicial set of dimension 1 (so no nondegenerate simplices above dimension 1), if we try to mimic the construction of the incidence algebra for a simplicial complex, $K$, one obvious way  is to give a total order to the vertices of $K$ and, denoting the set of (non-degenerate) $n$-simples in $K$ by $K_n$, to form $C(K)_n$ a vector space with basis $\{e_\sigma ~|~ \sigma \in K_n\}$, or abusing notation $\{e_\sigma ~|~ \sigma \in K_n\}$.  The face maps $d_i : K_n \to K_{n-1}$, which are defined since we have ordered the vertices, induce linear maps, $d_i : C(K)_n \to C(K)_{n-1}$ and yield a structure that is almost a simplicial vector space. (It only fails because of the lack of degeneracies.)  The usual construction on such a gadget is to take the alternating sum
$$\delta = \sum (-1)^id_i : C(K)_n \to C(K)_{n-1}$$and then a routine calculation shows that $\delta\delta = 0$, so $(C(K)_\bullet, \delta)$ is a chain complex.

\textbf{Remark}

The algebraic topology of $(C(K),\delta)$ iswell known.   The homology of this chain complex is the homology of the original simplicial complex, $K$ a construction that was at the heart of the construction of simplicial homology in the early 20th century.  The dual cochain complex obtained by dualising all the spaces and boundary maps, $(C^*(K),\delta^*)$,  gives the simplicial cohomology.  We refer to  \cite{kar2}  and \cite{lehmann} for a discussion of the problem of giving this cochain complex  a commutative multiplication.  The usual method (Alexander-Whitney) is not commutative. For us, that is not that important, but we do need to adjust its structure to bring it nearer to the de Rham complex.

This chain complex has additional structure as the \emph{geometry} of $K$ has been encoded in the given basis, $\{e_\sigma ~|~ \sigma \in K_n\}$ for each $C(K)_i$.  Assume that $K_i$ is finite and use the given basis to identify $C(K)_i$ and its dual $C(K)^*_i$ via the obvious innerproduct  
$$\langle \quad | \quad \rangle : C(K)_i\times C(K)_i \to k \enspace ,$$
for $k$ a field.
In \cite{zapatrin}, Zapatrin replaces the cochain complex differential $\delta^T$, obtained by taking the transpose of $\delta$ by   a new differential given by : \\
if $x_0\leq x_1\leq \ldots \leq x_n$ is an $n$-simplex,\\
$d(x_0\leq x_1\leq \ldots \leq x_n)=  \sum_{y<x_0}y\leq x_0\leq x_1\leq \ldots \leq x_n$\\
\hspace*{.5cm} $+\sum_{m=1}^n (-1)^m \sum_{ x_{m-1}<y<x_m}x_0\leq x_1\leq \ldots\leq x_{m-1} \leq y \leq x_m \leq \ldots \leq x_n$\\
\hspace*{1.5cm}$+ (-1)^{n+1} \sum_{x_n<y} x_0\leq x_1\leq \ldots \leq x_n\leq y$\enspace .\\
In fact, he restricts to strictly increasing chains having $<$ instead of $\leq$. We will discuss the precise relationship of this with $\delta^T$ in more detail in the second part of this paper, which will be, in part, devoted to an analysis of what information this Zapatrin model tells one about the simplicial complex, $K$.

\section{Vietoris-de Rham theorems}

As explained above, one possible test for a `good' theory of differential forms on a class of spaces, might  be to see if it satisfied a version of the de Rham theorem linking de Rham cohomology (relative to that theory) with a topologically based theory.  The following sketches out a general attack on this, based on an idea used by Allday and Halperin, \cite{CA-SH}.

Suppose that we have a functorial construction  $A_{dR}$ from the opposite of category of simplicial sets to that of dgas, i.e. differential graded algebras.  Many of those that we have considered are such that a form of de Rham theorem links the cohomology of $A_{dR}(K) $ with the ordinary cohomology of $K$ with real coefficients, so we will assume that this holds for ourfavouritee one of the moment, fixing $k = \mathbb{R}$ in the process.

Define for any space $X$, a direct system of dgas  by $$A(X) = \{A_{dR}(V(\mathcal{U})) ~|~ \mathcal{U} \textrm{ an open cover of } X\}\enspace ,$$
the `bonding' maps in this direct system being induced by those of the Vietoris construction itself. Now let $\overline{A}(X)$ be the colimit (direct limit) of 
$A(X)$.  This is an algebra of differential forms and we might call it the Vietoris-de Rham algebra of $X$. Then a version of the de Rham theorem follows for this complex.  The argument goes that, as homology commutes with direct limits, 
$$H(\overline{A}(X))  \cong colim H(A_{dR}(V(\mathcal{U}))\enspace .$$
The assumption that a de Rham theorem follows holds for $A_{dR}$, means that this is naturally isomorphic to $colim H(V(\mathcal{U}), \mathbb{R}).$  However as is known classically,  this latter cohomology is precisely the Alexander-Spanier cohomology of $X$ with real coefficients.  

Note that nowhere did we have to use any special features about the space $X$, so what is the problem?  

\emph{The most computational constructive and geometric generalisations of the de Rham complex are best behaved only on finite simplicial complexes and $V(\mathcal{U})$ is not finite.}

\noindent We could replace $V(\mathcal{U})$  by the Sorkin model relative to the cover $\mathcal{U}$, and the same argument works.  The resulting cohomology does not seem to change since the \v{C}ech and Vietoris models are homotopically equivalent, by Dowker's result, so \v{C}ech based cohomology and Alexander-Spanier
cohomology are naturally ismorphic. 

Another problem is to link up the Zapatrin model with all this.  The assumptions we made are not all known for that model, but it does give, as we will see in the second half of this paper, a good interpretation of the geometry of the complex and in certain cases (perhaps for all fractafolds) the representation of that geometry  seems to pass to the limiting dga.

If one tries to use the \v{C}ech construction to replace the Vietoris one, and hence to avoid `points' then problems arise with the refinement maps as $A(N(\mathcal{U}))$ will not be a directed system of dgas, merely one `up to coherent homotopy'.  However, for our supposed class of fractafolds, it seems likely that each will be represented by an approximating sequence of polyhedra with excellent control over the refinement maps `locally' and one can use that `excellent behaviour' to define a limiting algebra of differential forms in a highly controlled way.  At this point however we are getting beyond our present knowledge and into a more speculative range of questions, so in the second part of this paper, we will look at limiting processes, recalling known aspects of them for the situation of the Sorkin model and then will look at how the Zapatrin-de Rham model behaves for simple examples and at its more detailed algebraic structure.

\bibliographystyle{splncs}

\end{document}